%% file: main.tex
\documentclass[letterpaper]{article} 
\usepackage{aaai2026} 
\usepackage{times}  
\usepackage{helvet}  
\usepackage{courier}  
\usepackage[hyphens]{url}  
\usepackage{graphicx} 
\usepackage{natbib}  
\usepackage{caption} 
\usepackage{algorithm}
\usepackage{algorithmic}
\usepackage{multirow}
\usepackage{booktabs}
\usepackage[table]{xcolor}
\usepackage{amsmath,amssymb}
\usepackage{pifont}
\usepackage{newfloat}
\usepackage{listings}
\usepackage{pifont}
\DeclareCaptionStyle{ruled}{labelfont=normalfont,labelsep=colon,strut=off} 
\floatstyle{ruled}
\newfloat{listing}{tb}{lst}{}
\floatname{listing}{Listing}

\nocopyright 

\title{Uncertainty-Aware GUI Agent: Adaptive Perception through Component Recommendation and Human-in-the-Loop Refinement}
\author {
    Chao Hao\textsuperscript{\rm 1},
    Shuai Wang\textsuperscript{\rm 2}\thanks{Corresponding Author},
    Kaiwen Zhou\textsuperscript{\rm 3}
}
\affiliations {
    \textsuperscript{\rm 1} The Hong Kong
Polytechnic University\\
    \textsuperscript{\rm 2} Peking University\\
    \textsuperscript{\rm 3}The Chinese University of Hong Kong\\
    fanye-chao.hao@connect.polyu.hk,
    wangshuai\_2016@pku.edu.cn, 
    kwzhou@cse.cuhk.edu.hk
}

\usepackage{bibentry}

\begin{document}

\maketitle

\begin{abstract}
Graphical user interface (GUI) agents have shown promise in automating mobile tasks but still struggle with input redundancy and decision ambiguity. In this paper, we present \textbf{RecAgent}, an uncertainty-aware agent that addresses these issues through adaptive perception. We distinguish two types of uncertainty in GUI navigation: (1) perceptual uncertainty, caused by input redundancy and noise from comprehensive screen information, and (2) decision uncertainty, arising from ambiguous tasks and complex reasoning. To reduce perceptual uncertainty, RecAgent employs a component recommendation mechanism that identifies and focuses on the most relevant UI elements. For decision uncertainty, it uses an interactive module to request user feedback in ambiguous situations, enabling intent-aware decisions. These components are integrated into a unified framework that proactively reduces input complexity and reacts to high-uncertainty cases via human-in-the-loop refinement. Additionally, we propose a dataset called \textbf{ComplexAction} to evaluate the success rate of GUI agents in executing specified single-step actions within complex scenarios. Extensive experiments validate the effectiveness of our approach. The dataset and code will be available at \url{https://github.com/Fanye12/RecAgent}.


\end{abstract}

\begin{figure}[h]
      \centering
      \includegraphics[width=0.95\linewidth]{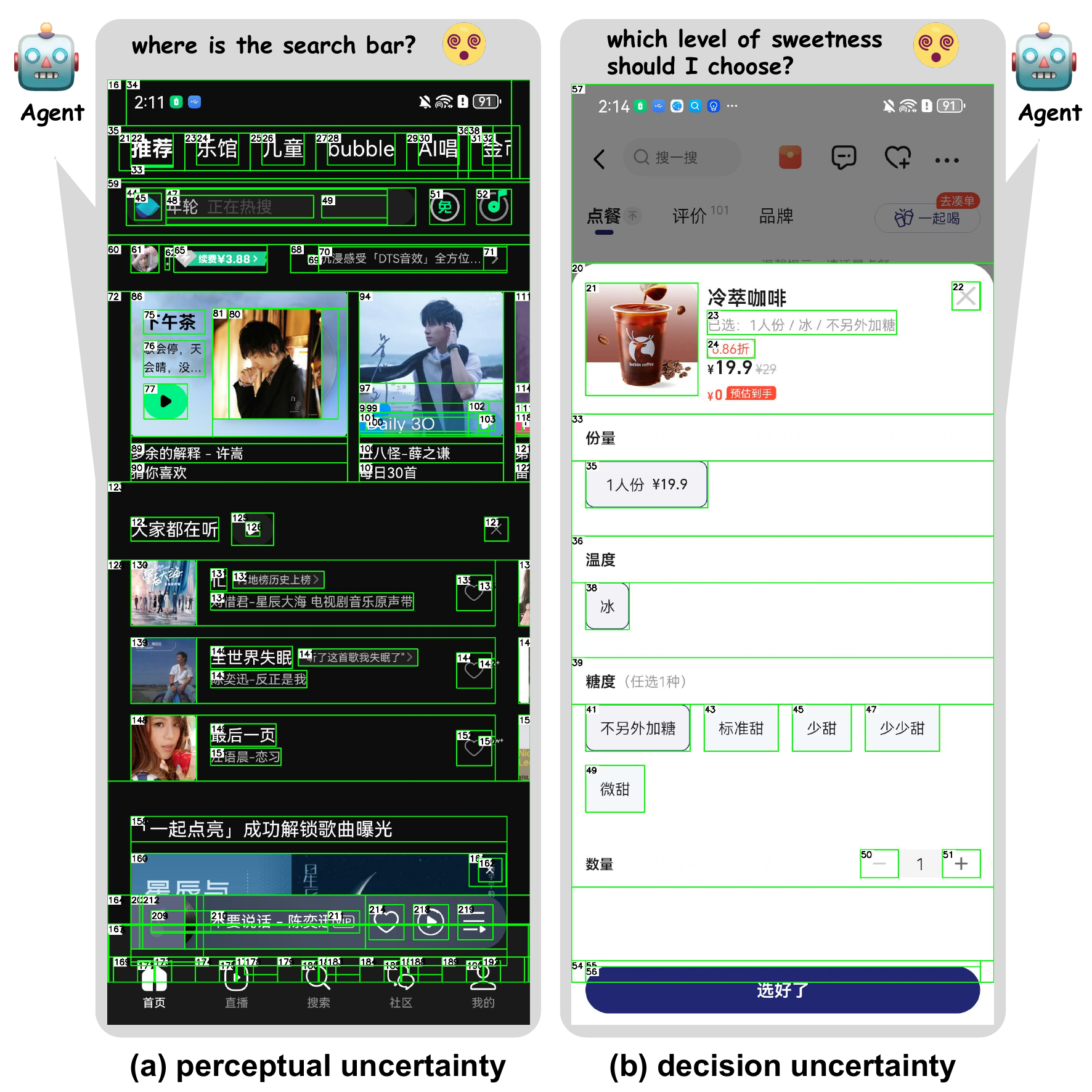}
      \vspace{-2mm}
      \caption{The two major challenges faced by existing GUI Agents: (a) Perceptual uncertainty caused by input redundancy. For example, when searching in a music application, excessive input redundancy prevents the Agent from locating the search box component. (b) Decision uncertainty caused by the lack of interactive mechanisms. For example, the Agent does not know which sweetness level to select for the user when helping with coffee ordering. The input used here is in SoM \cite{som} format.}
      \label{fig:motivation}
      \vspace{-5mm}
\end{figure}

\section{Introduction}
\label{sec:intro}

Graphical user interface (GUI) agents aim to automate human interactions with mobile applications, enabling users to accomplish tasks such as ordering food, checking the weather, or booking tickets by simply specifying high-level goals \cite{seeact, androidworld}. While recent progress in large language models (LLMs) \cite{gpt4, gemini1.5, Qwen-VL} and vision-language systems has empowered GUI agents with improved planning and action capabilities \cite{survey1, survey2, survey3}, several core challenges remain unresolved. In particular, two critical issues hinder their reliability and generalizability in complex real-world applications: \textbf{input redundancy} and \textbf{decision ambiguity}, as shown in Figure~\ref{fig:motivation}.

Many existing GUI agents take a comprehensive approach to perception, incorporating both full-screen screenshots and complete UI element lists as input \cite{androidworld, GUI-explorer, mobileagentv2}. While this exhaustive strategy ensures access to all available information, it also introduces significant noise and redundancy. This input redundancy has two clear drawbacks: on one hand, it leads to low computational efficiency; on the other hand, it interferes with the model's ability to perceive truly useful information \cite{efficient, efficient2, efficient3, langtime, xufault}. For instance, dozens or even hundreds of UI elements may be irrelevant to the current subtask, overwhelming the agent and complicating both perception and reasoning. We define this problem as \textit{perceptual uncertainty}, an uncertainty arising from the agent's inability to focus on truly relevant components due to overloaded input. As shown in Figure~\ref{fig:motivation}(a), due to excessive input redundancy, the Agent cannot locate the desired search box to complete the search action.

Another major challenge is the agent's inability to handle \textit{decision uncertainty}, especially in ambiguous scenarios that involve user preferences or require disambiguation. For example, as shown in Figure~\ref{fig:motivation}(b), when ordering coffee, users often provide only rough instructions such as “help me order a coffee”. However, the actual operation typically requires selecting the user's desired sweetness or temperature. Most existing systems adopt a purely autonomous approach and lack mechanisms to interactively solicit user feedback during execution \cite{androidworld, appagent}, leading to unsatisfactory or failed actions.

To address these two challenges, we propose \textbf{RecAgent}, an uncertainty-aware GUI agent that enhances both perception and decision-making through adaptive mechanisms. Specipically, RecAgent is composed of several functional agents: \textit{Planning Agent}, \textit{Decision Agent}, \textit{Reflection Agent}, and \textit{Interaction Agent}, alongside two auxiliary modules: a \textit{Component Recommendation Module} and a \textit{Memory Unit}.
The overall process starts with the Planning Agent, which generates the current subgoal based on the user task and the observed environment state. To combat perceptual uncertainty, we introduce a recommendation-based perception mechanism: the Component Recommendation Module selectively filters and ranks relevant UI elements from the environment, using keyword matching, semantic similarity, and historical context. Instead of providing the Decision Agent with an entire UI tree, only the top-ranked elements (e.g., top $10$) are passed forward, substantially reducing input size while preserving essential information.

Next, the Decision Agent takes the current subgoal and the filtered UI components to predict the optimal action. This action is executed in the environment, resulting in a new state. The Reflection Agent evaluates whether the current subgoal has been successfully completed. If the action fails, the agent rolls back, excludes the previous UI element choice, and attempts alternative actions until success.

In cases where the system encounters high decision uncertainty (e.g., multiple valid options or missing preferences), the Interaction Agent determines whether user input is required. If necessary, it dynamically generates a query to the user (e.g., “What level of sweetness do you prefer? ”), and integrates the feedback into the ongoing execution. All intermediate observations and decisions are recorded and updated in the Memory Module for continual learning and future planning.

Furthermore, we introduce a new evaluation dataset, \textbf{ComplexAction}, designed to assess the agent's capability to execute fine-grained single-step actions within visually and semantically complex environments. Unlike previous benchmarks that focus solely on end-to-end task completion \cite{aguvis, mirage, spabench, chop}, ComplexAction focuses exclusively on the success rate of executing a specified single-step action within complex scenarios, thereby providing a better validation of the effectiveness of our component recommendation module.

In summary, our contributions are threefold:
\begin{itemize}
    \item We identify and tackle two forms of uncertainty in GUI agents: \textit{perceptual} and \textit{decision} uncertainty, that hinder performance in real-world applications.
    \item We propose RecAgent, a novel uncertainty-aware GUI agent featuring a recommendation-based perception mechanism and a human-in-the-loop interaction module.
    \item We construct ComplexAction, a new benchmark dataset for evaluating single-step GUI action accuracy in complex mobile interface scenarios.
\end{itemize}

Extensive experiments demonstrate that RecAgent outperforms existing baselines in both overall success rate and step-wise accuracy, especially in complex environments with high uncertainty.

\section{Related Work}
\label{sec:related}

\textbf{GUI Agents}. Recent advancements in GUI agents have largely leveraged the powerful multimodal understanding capabilities of Multimodal Large Language Models (MLLMs) and vision-language models \cite{gpt4, gemini1.5, qwenvl2.5}. Representative works include the AppAgent series \cite{appagent, appagentv2, appagentx}, the Mobile-Agent series \cite{mobileagent, mobileagentv2, mobileagente}, and so on. Early systems like AppAgent \cite{appagent} and AutoDroid \cite{autodroid} demonstrated task automation using foundation models. The integration of visual perception with LLMs has been explored in systems such as SeeAct \cite{seeact}, which utilizes GPT-4V for web task automation, and MobileAgentV2 \cite{mobileagentv2}, which employs a multi-agent architecture with memory for mobile tasks. M3A \cite{androidworld} combines ReAct-style reasoning with Set-of-Mark \cite{som} visual annotations for Android control, showcasing strong generalization.

\vspace{2mm}
\noindent\textbf{Addressing Input Redundancy}:
The problem of input redundancy, which we identify as a source of \textit{perceptual uncertainty}, has garnered attention. Some work focuses on improving efficiency through better action grounding or hierarchical perception \cite{efficient, Simplification}. Techniques like the Set-of-Mark (SoM) representation \cite{som} aim to simplify the input space by annotating key elements. Our approach differs by introducing an adaptive, recommendation-based mechanism to actively filter and prioritize relevant UI components, directly mitigating the negative impact of overloaded inputs on perception and reasoning.

\vspace{2mm}
\noindent\textbf{Handling Decision Ambiguity}:
Decision ambiguity, or \textit{decision uncertainty}, particularly in user preference elicitation, is another recognized challenge. Interactive frameworks have been proposed where agents ask questions to resolve ambiguities, such as those for mobile navigation \cite{autoglm} or general communication \cite{seed}. However, integrating such interactive capabilities seamlessly into the GUI agent's execution loop, especially for mobile tasks, remains less explored. Our work proposes a dedicated interaction module within RecAgent that dynamically identifies high-uncertainty states and solicits user feedback, enabling intent-aware decision-making in ambiguous scenarios like customization choices.

\begin{figure}[t]
      \centering
      \includegraphics[width=0.95\linewidth]{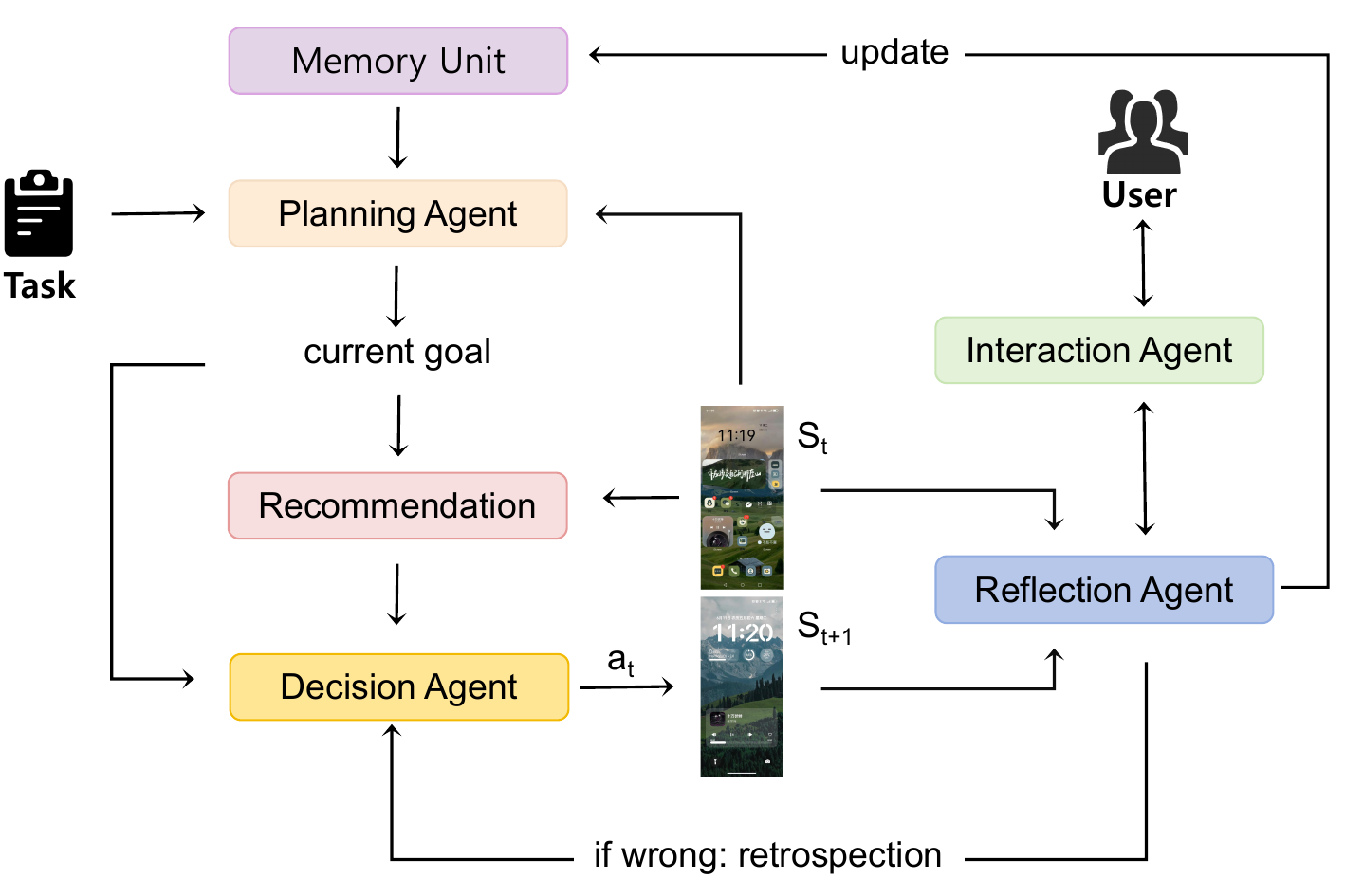}
      \vspace{-3mm}
      \caption{Schematic overview of the RecAgent architecture.}
      \label{fig:framework}
      \vspace{-3mm}
\end{figure}

\begin{figure}[t]
      \centering
      \includegraphics[width=0.95\linewidth]{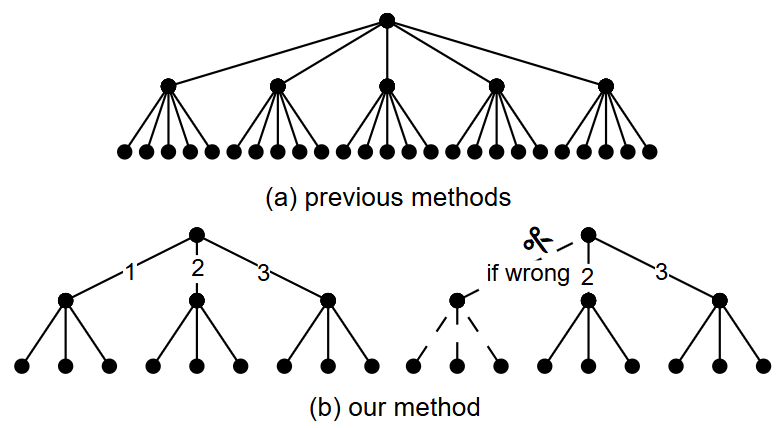}
      \vspace{-3mm}
      \caption{Comparison with previous methods. Schematic diagram of using the component recommendation module in conjunction with the retrospection mechanism. The recommendation module reduces the complexity of the input and narrows down the path choices. When it is determined that a previous action was ineffective, the retrospection mechanism deletes the previously chosen path and reselects a possible path (i.e., action).}
      \label{fig:retrospection}
      \vspace{-5mm}
\end{figure}

\section{Method}

In this section, we present the architecture and workflow of \textbf{RecAgent}, an uncertainty-aware GUI agent designed to reduce perceptual and decision uncertainty during task execution. The overall agent consists of four major components: \textit{Planning Agent}, \textit{Decision Agent}, \textit{Reflection Agent} and \textit{Interaction Agent}, along with two key modules: the \textit{Component Recommendation Module} and a shared \textit{Memory Unit} maintaining intermediate and historical information. The framework is illustrated in Figure~\ref{fig:framework}.

\subsection{Task Formalization}
Given a high-level task instruction $T$, RecAgent interacts with the mobile device environment $\mathcal{E}$ to generate a sequence of actions $\{a_1, a_2, \dots, a_n\}$ that completes the task. At each time step $t$, the agent observes the environment state $s_t$, makes an adaptive decision, and executes an action $a_t$ to reach the next state $s_{t+1}$. Through continuous interactions, we obtain a trajectory
$\mathcal{J} = {(s_1, a_1),(s_2, a_2),(s_3, a_3), . . . ,(s_L , a_L)}$, where $L$ represents the length of the trajectory. If the final state $s_L$ reaches the goal state, the task execution is considered successful.

\subsection{Task Decomposition via Planning Agent}

The Planning Agent is responsible for decomposing the overall task $T$ into a sequence of intermediate subgoals. At each step $t$, it generates the current subgoal $g_t$ based on the task $T$, current environment state $s_t$, and historical memory $M_{t-1}$:
\begin{equation}
    g_t = \text{Planner}(T, s_t, M_{t-1}).
\end{equation}
Here, $g_t$ represents the immediate semantic intent (e.g., ``click the search bar'', ``choose delivery method'') and serves as the guiding signal for perception and decision-making. This task decomposition enables the agent to gradually transform complex high-level tasks into executable concrete actions, thereby reducing the complexity and uncertainty in task execution, while also aligning with humans' step-by-step operational habits.

\subsection{Component Recommendation for Adaptive Perception}

The GUI environment state $s_t$ consists of a list of UI elements $\mathcal{U}_t = \{u_t^{(1)}, u_t^{(2)}, \dots, u_t^{(N)}\}$ and current screenshot $c_s$ (optional), where each $u_t^{(i)}$ is a structured representation containing text, bounds, type, etc. Some complex interfaces can have as many as hundreds of UI elements, processing all elements leads to redundant and noisy inputs \cite{mirage, Simplification}.

Inspired by the ``multi-channel recall'' mechanism in recommendation systems \cite{recommendation1, recommendation2}, we treat the current subgoal $g_t$ as a query and the UI elements in $\mathcal{U}_t$ as candidate items. To efficiently identify components most relevant to the current task intent, we design a component recommendation module (CRM) that employs multiple independent recommendation pathways. Each pathway generates a candidate subset in parallel based on distinct matching logic, and the final perception input is formed by taking the union of all outputs: ensuring high coverage while significantly reducing input scale.

Specifically, we define the following recommendation pathways:

\textbf{Keyword Matching Pathway}: Identifies keywords in $g_t$ (e.g., ``search'', ``submit'', ``open'') and matches them exactly or fuzzily against the text of UI elements;
    
\textbf{Semantic Matching Pathway}: Leverages a pretrained language model (e.g., BERT \cite{bert}) to assess the semantic relevance between $g_t$ and the text of each UI element, retrieving controls with implicit functional alignment;
    
\textbf{LLM-based Intent Recommendation Pathway}: Uses a LLM to perform contextual understanding of both $g_t$ and each $u_t^{(i)}$, determining whether their functional roles align, and outputs high-confidence recommendations.

Let $\mathcal{R}_k(g_t, \mathcal{U}_t)$ denote the candidate set returned by the $k$-th recommendation pathway. The final perception input set $\mathcal{U}_t^\prime$ is defined as the union of all pathway outputs:
\begin{equation}
    \mathcal{U}_t^\prime = \bigcup_{k} \mathcal{R}_k(g_t, \mathcal{U}_t).
\end{equation}

This set $\mathcal{U}_t^\prime$ includes any UI element deemed relevant by at least one pathway, thereby mitigating the risk of missed detections due to reliance on a single strategy, while effectively compressing the input space. Compared to processing all $N$ elements, this approach significantly reduces computational overhead and noise in the perception module, enhancing the system's robustness and response efficiency in complex GUI environments.

\subsection{Decision Making via Decision Agent}

Given the current subgoal $g_t$ generated by the Planning Agent and the filtered UI element list $\mathcal{U}_t^\prime$ produced by the Component Recommendation Module, the Decision Agent selects an executable action $a_t$ and generates a natural language description $d_t$ of the intended behavior:
\begin{equation}
    a_t, d_t = \text{Decision}(g_t, \mathcal{U}_t^\prime).
\end{equation}

The action space follows the M3A Agent \cite{androidworld} framework and includes a comprehensive set of GUI operations such as \texttt{click}, \texttt{double-click}, \texttt{text input}, and so on. $d_t$ provides a human-readable explanation (e.g., ``Click on the search bar to begin entering the query'').

By operating on the compact and semantically relevant subset $\mathcal{U}_t^\prime$, the Decision Agent reduces the search space for candidate targets, thereby mitigating decision uncertainty and improving both accuracy and efficiency. The inclusion of $d_t$ enables better interpretability and facilitates downstream components such as reflection and memory updating.

For actions that require interaction, such as \texttt{click}, each action is structured as a tuple $(\text{action\_type}, \text{target\_element})$, where $\text{target\_element} \in \mathcal{U}_t^\prime$ is the UI component selected for interaction. For predefined navigation actions such as \texttt{scroll up} or \texttt{navigate back}, the Operator directly invokes the corresponding system API without requiring precise element localization.

\subsection{Retrospection Mechanism via Reflection Agent}

After executing action $a_t$ and transitioning to the next state $s_{t+1}$, the Reflection Agent evaluates the outcome and produces both a success indicator and a contextual summary:
\begin{equation}
    (\text{Success}_t, \text{Summary}_t) = \text{Reflect}(g_t, s_t, s_{t+1}).
\end{equation}
Here, $\text{Success}_t \in \{\text{True}, \text{False}\}$ indicates whether the action $a_t$ successfully advanced progress toward subgoal $g_t$, while $\text{Summary}_t$ is a natural language or structured description summarizing the observed changes, such as ``The search bar was clicked, but no keyboard appeared'' or ``The page scrolled down, revealing more product items.''

We introduce a retrospection mechanism, illustrated in Figure~\ref{fig:retrospection}. If $\text{Success}_t$ is $\text{False}$, indicating that the action failed to achieve the intended effect, the agent reverts to the previous state $s_t$, removes the previously selected action from the filtered candidate set $\mathcal{U}_t^\prime$ and reinvokes the Decision Agent to select an alternative action. This backtracking process enables the agent to dynamically revise its decisions without restarting the entire task.

The tight integration between the retrospection mechanism and the Component Recommendation Module plays a critical role in enhancing system robustness. As shown in Figure~\ref{fig:retrospection}, this coordination reduces the effective state space by eliminating invalid candidates from future consideration, thereby avoiding repeated failures on the same element. Furthermore, it mitigates the impact of perceptual inaccuracies (e.g., mislocalized elements) or reasoning errors (e.g., incorrect action selection), allowing the agent to recover gracefully and explore alternative interaction paths. This closed-loop feedback design significantly improves the agent’s adaptability and reliability in complex and dynamic GUI environments.

\subsection{User Feedback via Interaction Agent}

To proactively manage \textit{decision uncertainty}, the Interaction Agent evaluates the need for user feedback after each successful action execution. Given the current subgoal $g_t$, the environment state $s_{t+1}$ (resulting from action $a_t$), and the action description $d_t$, the Interaction Agent determines whether user input is required:
\begin{equation}
    \text{need\_feedback}_t, q_t = \text{Interact}(g_t, a_t, s_{t+1}, d_t).
\end{equation}
Here, $\text{need\_feedback}_t \in \{\texttt{True}, \texttt{False}\}$ is a boolean flag indicating whether feedback is necessary. If $\text{need\_feedback}_t$ is $\texttt{True}$, $q_t$ is a natural language query string (e.g., ``what level of sweetness do you prefer?'') posed to the user. If $\text{need\_feedback}_t$ is $\texttt{False}$, $q_t$ is set to $\texttt{None}$.

If user feedback is required ($\text{need\_feedback}_t$ is $\texttt{True}$), the agent awaits the user's response $u_t$. This response is then incorporated to refine the next subgoal:
\begin{equation}
    g_{t+1}^\prime = \text{UpdateGoal}(g_{t+1}, u_t).
\end{equation}
The agent then proceeds with subsequent steps using the (potentially updated) subgoal $g_{t+1}$. This mechanism allows RecAgent to resolve ambiguities and align its actions with user intent in real-time.

\subsection{Memory Unit and Task Termination}

Throughout the interaction process, all subgoals, actions, description, success indicators, summary, query and user feedback (if any) are stored in the Memory Unit:
\begin{equation}
    M_t = M_{t-1} \cup \{(g_t, a_t, d_t, \text{Success}_t, \text{Summary}_t, q_t, u_t)\}.
\end{equation}

The task terminates when either:
\begin{itemize}
    \item The Decision Agent outputs a special \texttt{COMPLETE} action indicating that the task has been completed, or
    \item The agent reaches the maximum number of allowed steps $L_\text{max}$.
\end{itemize}

\subsection{Overall Algorithm}

The full execution loop is summarized in Algorithm~\ref{alg:recagent}, combining planning, recommendation, decision, reflection, and interaction in an uncertainty-aware loop.

\begin{algorithm}[h]
\caption{RecAgent Execution Loop}
\label{alg:recagent}
\begin{algorithmic}[1]
\STATE \textbf{Input:} Task $T$, initial state $s_0$
\STATE Initialize memory $M_0 = \emptyset$
\STATE $s \gets s_0$
\FOR{$t = 1$ to $L_{\max}$}
    \STATE $g_t \gets \text{Planner}(T, s, M_{t-1})$
    \STATE \textbf{// Component Recommendation}
    \STATE $\mathcal{U}_t^\prime \gets \bigcup_{k} \mathcal{R}_k(g_t, \mathcal{U}_t)$
    \STATE \textbf{// Decision Making}
    \STATE $(a_t, d_t) \gets \text{Decision}(g_t, \mathcal{U}_t^\prime)$
    \STATE Execute $a_t$, observe new state $s'$
    \STATE \textbf{// Reflection}
    \STATE $(\text{Success}_t, \text{Summary}_t) \gets \text{Reflect}(g_t, s, s')$
    \IF{$\text{Success}_t == \texttt{False}$}
        \STATE Remove selected element from $\mathcal{U}_t^\prime$
        \STATE Reinvoke Decision Agent with updated $\mathcal{U}_t^\prime$
        \STATE Continue
    \ENDIF
    \STATE \textbf{// Interaction}
    \STATE $(\text{need\_feedback}_t, q_t) \gets \text{Interact}(g_t, s', d_t)$
    \IF{$\text{need\_feedback}_t == \texttt{True}$}
        \STATE Receive user input $u_t$
        \STATE $g_{t+1} \gets \text{UpdateGoal}(g_{t+1}, u_t)$
    \ELSE
        \STATE $u_t \gets \texttt{None}$
    \ENDIF
    \STATE \textbf{// Memory Update}
    \STATE $M_t \gets M_{t-1} \cup \{(g_t, a_t, d_t, \text{Success}_t, \text{Summary}_t, q_t, u_t)\}$
    \STATE $s \gets s'$
    \IF{$a_t == \texttt{[COMPLETE]}$}
        \STATE \textbf{break}
    \ENDIF
\ENDFOR
\end{algorithmic}
\end{algorithm}

\section{The ComplexAction Dataset}
\label{sec:dataset}

To effectively evaluate the capability of GUI agents in handling \textit{perceptual uncertainty} within complex environments and to specifically validate the effectiveness of our Component Recommendation Module, we introduce the ComplexAction dataset.

\vspace{2mm}
\noindent\textbf{Motivation and Design Principles.}
Unlike existing benchmarks that primarily focus on end-to-end task completion rates, the ComplexAction Dataset is designed to assess an agent's accuracy in executing \textit{fine-grained, single-step actions} (e.g., clicking a specific button) within visually and semantically complex GUI scenes. This focus allows for a more precise measurement of an agent's core perception and decision-making abilities, particularly its ability to locate relevant UI elements amidst significant input redundancy. Our dataset directly targets the evaluation of mechanisms designed to mitigate perceptual uncertainty.

\vspace{2mm}
\noindent\textbf{Data Collection and Structure.}
We identify five common and representative action types: \textit{Click Search Box}, \textit{Create New Content}, \textit{Like/Upvote}, \textit{Refresh Interface}, and \textit{Sort Items}. For each action type, we collect scenarios from popular mobile applications in China (e.g., Pinduoduo, Tencent Video, Xiaohongshu, etc.), resulting in a total of $62$ diverse and complex scenes. A scene is considered ``complex" if its GUI state contains hundreds of UI elements, presenting a significant challenge for agents that rely on full-state inputs. Figure~\ref{fig:motivation}(a) provides an example of such a scenario.

For each collected scene, we provide the raw screen screenshot and the parsed list of UI elements as input. The ground truth is defined by the specific target UI element for the designated action. Evaluation can be performed by checking if the agent's predicted action targets the correct element or by verifying if the subsequent UI state transition aligns with the expected outcome on a real device, as judged by human annotators.

This dataset facilitates a targeted evaluation of an agent's ability to focus on relevant components, thereby providing a more granular assessment of its robustness in challenging perceptual conditions. More details can be found in the appendix.

\section{Experiments}
\input{table/androidworld}

\input{table/miniwob}

\input{table/complexaction}

\subsection{Experimental Setting}

\subsubsection{Implementation Details}
 We build RecAgent based on the M3A Agent \cite{androidworld}, adopting the same action space. For the newly introduced modules, we design the prompts in a style consistent with M3A. Additionally, similar to M3A, we also implement two input modalities: accessibility tree (a11y tree) and SoM (a11y tree + screenshot) \cite{som}. Following most work in the field \cite{GUI-explorer, mirage}, we use the widely adopted GPT-4o as the base model. In the semantic matching pathway, we use the text-embedding-3-small model to compute text similarity. The maximum number of steps per task is set to $30$.

\subsubsection{Datasets}
We evaluate the proposed RecAgent against several related works on the following datasets:

\textbf{AndroidWorld}:
 AndroidWorld is an Android environment that includes 116 tasks drawn from 20 real-world applications. The benchmark dynamically generates task variants by randomizing parameters such as message content, contact names, and calendar dates, resulting in millions of unique task instances.
 
\textbf{MobileMiniWoB++}: MobileMiniWoB++  \cite{androidworld} adapts the MiniWoB++ \cite{miniwob1, miniwob2} benchmark to the Android environment within the AndroidWorld framework. It renders web tasks using native UI components, and uses accessibility trees and screenshots for observation, supporting 92 tasks after compatibility filtering.

\textbf{ComplexAction}: The dataset proposed in this paper for evaluating the success rate of given single-step actions in complex scenarios is also built upon the AndroidWorld framework, with a detailed introduction provided earlier.

Note that the evaluation of RecAgent using these datasets does not require the involvement of the Interaction Agent.

\subsubsection{Comparative Baselines} 
We select several SOTA methods in the field for performance comparison. Relevant methods compared include SeeAct \cite{seeact}, M3A/T3A \cite{androidworld}, MobileAgentV2 \cite{mobileagentv2} etc., among which M3A is the most closely related to our approach. Additionally, we evaluate approaches that solely use screenshots as input, such as Mirage \cite{mirage} and CogAgent \cite{cogagent}.

\subsection{Comparision With SOTA Methods}
As shown in Table~\ref{tab:aw}, RecAgent demonstrates outstanding performance on the AndroidWorld benchmark, with its SoM-based input variant achieving the best results among all compared methods. Compared to the most relevant baseline methods M3A and T3A, it achieves performance improvements of $17.6\%$ and $15.7\%$ under two input conditions respectively, which fully validates the effectiveness of our proposed approach.

Notably, while both GUI-Explorer and Mirage also deliver strong performance, their methods require either pre-exploration or iterative knowledge accumulation in relevant environments due to their exploration and knowledge extraction capabilities. In contrast, our RecAgent can be deployed directly and exhibits superior generalization ability.

Similarly, the proposed RecAgent's SoM-based input variant achieves optimal performance on the MobileMiniWoB++ dataset, as shown in Table~\ref{tab:mw}. While all methods demonstrate decent performance on this relatively simpler dataset (with human achieving $100\%$ accuracy), our RecAgent still attains the best performance at $69.8\%$, further validating RecAgent's generalization capability. Since the scenarios in this dataset are relatively simple, our method shows only marginal improvements over M3A and T3A.

As shown in Table~\ref{tab:ca}, RecAgent demonstrates strong performance on the ComplexAction dataset. Specifically, the SoM-based input variant achieves a success rate of $69.3\%$, outperforming all other methods except human-level performance ($100\%$). This result highlights the effectiveness of our approach in handling complex scenarios.

Compared to the most relevant baseline methods (M3A and T3A), RecAgent shows significant improvements in action success rates. While M3A achieves $64.5\%$ and T3A achieves $59.7\%$, RecAgent's SoM-based variant surpasses both by a notable margin. This further validates the robustness and generalization capability of RecAgent in challenging environments.


\begin{figure}[t]
      \centering
      \includegraphics[width=0.9\linewidth]{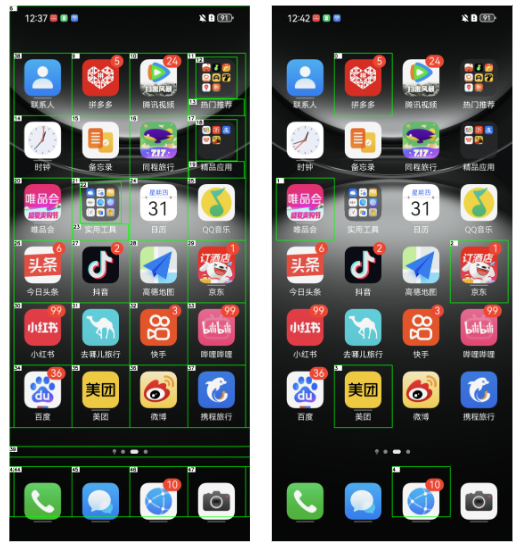}
      \vspace{-3mm}
      \caption{Visualization of SoM outputs before and after using the Component Recommendation Module. Left: original; right: with module. Current goal: ``open a shopping app”.}
      \label{fig:crm}
      \vspace{-4mm}
\end{figure}

\begin{figure}[t]
      \centering
      \includegraphics[width=0.9\linewidth]{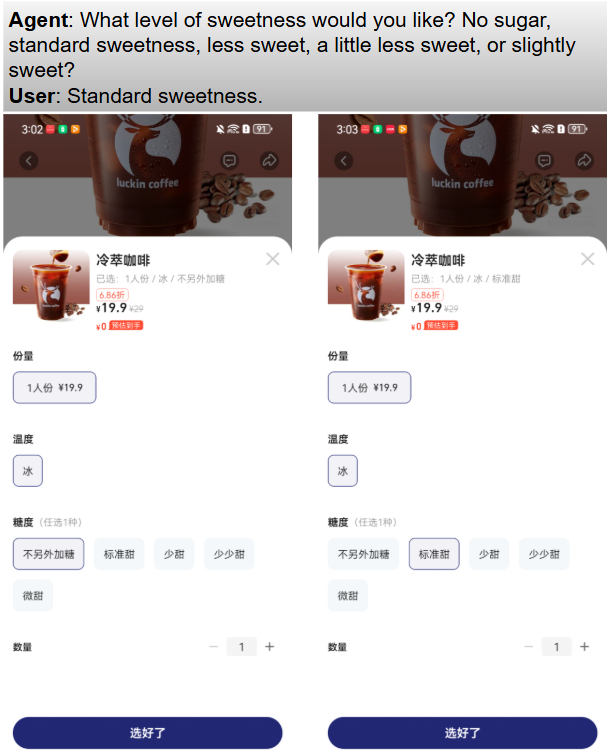}
      \vspace{-3mm}
      \caption{Visualization of the Interaction Agent. The left image shows a coffee-ordering scenario, where the interface presents multiple sweetness options. In this case, the agent proactively asks the user for their preference and, based on the user's response, selects the desired sweetness level, as shown in the right image.}
      \label{fig:interaction}
      \vspace{-5mm}
\end{figure}

\subsection{Qualitative Visualization}
\subsubsection{Reducing Perceptual Uncertainty} In Figure~\ref{fig:crm}, we present the visualization of the component recommendation module. It is clearly evident that, without the component recommendation module, all UI elements are annotated using the SoM \cite{som} method and displayed with numbered bounding boxes, regardless of the number of elements. In practice, lists of UI elements in the form of text are also included as input, leading to substantial input redundancy. When the component recommendation module is applied, only the most relevant UI elements are dynamically retained. In this example, where the goal is to open a shopping application, the preserved elements are the most popular shopping apps in China, such as Pinduoduo and Jingdong. Irrelevant text-based UI element lists are filtered out before input; in this case, the number of UI elements is reduced from 47 to 5. In more complex scenarios, the initial count can even reach hundreds. This significantly reduces input redundancy, thereby decreasing the agent's perceptual uncertainty and enabling more accurate localization of the target UI element. 

\subsubsection{Reducing Decision Uncertainty} 
In Figure~\ref{fig:interaction}, we present the visualization of the Interaction Agent. It can be clearly seen that in scenarios with ambiguous user intent, such as ordering coffee, the interaction agent proactively inquires about the user's preferences and makes appropriate selections based on the user's responses. This interactive mechanism is absent in most existing methods, and it helps reduce decision uncertainty, thereby better aligning with real-world user needs. Without this interactive mechanism, the agent would have to make decisions randomly or rely on default settings, making it difficult to achieve user-satisfying results.

\input{table/ablation1}
\input{table/ablation2}
\subsection{Ablation Study} 

\textbf{Effectiveness of component recommendation module and retrospection mechanism}. We present a schematic diagram in Figure~\ref{fig:retrospection} demonstrating the effect of using both the component recommendation module and the retrospection mechanism, which significantly reduces the complexity of the path space. We conducted ablation experiments in Table~\ref{tab:ablation1} to quantitatively analyze its effectiveness. It can be seen that using either part alone achieves some performance improvement, but the gains are not very significant. However, when the two are used together, they achieve the best results. Compared to the M3A baseline, our major improvements lie in CRM (Component Recommendation Module) and RM (Retrospection Mechanism), as demonstrated by the ablation experiments, which validate the effectiveness of the method we propose.

\noindent\textbf{Effectiveness of different recommendation pathways}. As shown in Table~\ref{tab:ablation2}, we present quantitative results using different recommendation pathways. It can be observed that when only KMP or SMP is used, performance decreases, as they cannot guarantee accurate recall of the required components. Using only LRP leads to a certain improvement in performance, but the gain is not significant. Only when all three are used together can the best performance be achieved.


\section{Conclusion}
In this paper, we present RecAgent, an uncertainty-aware GUI agent that addresses input redundancy and decision ambiguity in mobile task automation. RecAgent reduces perceptual uncertainty through a component recommendation mechanism that selectively focuses on relevant UI elements. To handle decision uncertainty, it incorporates an interactive module that seeks user feedback in ambiguous situations. These components are integrated into a unified framework that proactively simplifies inputs and reactively resolves uncertainties. Furthermore, we introduce the ComplexAction dataset to evaluate the success rate of agents in executing specific actions within complex scenarios. Extensive experiments demonstrate the effectiveness of our proposed method.

\label{sec:reference_examples}

\nobibliography*

\bibliography{aaai2026}

\end{document}

%% file: table/androidworld.tex
\begin{table}[t]
\centering
\resizebox{\columnwidth}{!}{
\begin{tabular}{lllc}
\toprule[1.5pt]
\textbf{Agent}         & \textbf{Input}              & \textbf{Base Model} & {\begin{tabular}[c]{@{}c@{}}\textbf{Task Success} \\ \textbf{Rate (\%)}\end{tabular}} \\
\midrule
\textit{Human~\citep{androidworld}}            & \textit{screen}             & \textit{-}     & \textit{80.0} \\

Aguvis \cite{aguvis}              & screen             & GPT-4o     & 37.1 \\
AppAgent \cite{appagent}            & SoM             & GPT-4o     & 14.9 \\
Aria-UI \cite{ariaui}             & screen             & GPT-4o     & 44.8 \\
AutoDroid \cite{autodroid}     & a11y tree              & GPT-4o     & 15.7 \\
T3A \cite{androidworld}                  & a11y tree              & GPT-4o & 37.6 \\
M3A \cite{androidworld}                 & SoM   & GPT-4o     & 40.5 \\
Ponder \& Press \cite{ponderpress}     & screen             & GPT-4o     & 34.5 \\
SeeAct \cite{seeact}               & SoM   & GPT-4-turbo & 15.5 \\
UGround \cite{uground}             & screen             & GPT-4o     & 32.8 \\
Mirage-O \cite{mirage} &screen &GPT-4o &42.2 \\
GUI-explorer \cite{GUI-explorer} & SoM & GPT-4o     & 47.4 \\
\rowcolor[HTML]{DDDDDD}
RecAgen (Ours)   &a11y tree &GPT-4o &43.5\\ 
\rowcolor[HTML]{DDDDDD}
RecAgen (Ours)   &SoM &GPT-4o &\textbf{47.8} \\ 
\bottomrule[1.5pt]
\end{tabular}}
\vspace{-2mm}

\caption{Performance comparison on AndroidWorld. Best results are in \textbf{bold}.}

\label{tab:aw}
\vspace{-3mm}

\end{table}

%% file: table/miniwob.tex
\begin{table}[t]
\centering

\resizebox{\columnwidth}{!}{
\begin{tabular}{lllc}
\toprule[1.5pt]
\textbf{Agent}         & \textbf{Input}              & \textbf{Base Model} & {\begin{tabular}[c]{@{}c@{}}\textbf{Task Success} \\ \textbf{Rate (\%)}\end{tabular}} \\
\midrule
\textit{Human~\citep{androidworld}}            & \textit{screen}             & \textit{-}     & \textit{100} \\

SeeAct \cite{seeact}               & SoM   & GPT-4 Turbo & 66.1 \\
AppAgent \cite{appagent}               & SoM   & GPT-4o & 56.1 \\
T3A \cite{androidworld}              & a11y tree             & GPT-4o     & 68.1 \\
M3A \cite{androidworld}            & SoM             & GPT-4o     & 68.5 \\
OS-Atlas* \cite{osatlas}             & screen             & GPT-4o     & 51.1 \\
UGround* \cite{uground}     & screen             & GPT-4o     & 48.4 \\
Mirage-O \cite{mirage}                  & screen              & GPT-4o & 60.9 \\
\rowcolor[HTML]{DDDDDD}
RecAgen (Ours)   &a11y tree &GPT-4o &68.8\\ 
\rowcolor[HTML]{DDDDDD}
RecAgen (Ours)   &SoM &GPT-4o &\textbf{69.8} \\ 
\bottomrule[1.5pt]
\end{tabular}
}
\vspace{-2mm}
\caption{Performance comparison on MobileMiniWoB++. * means
the results are reproduced under the same prompt setting.}

\label{tab:mw}
\vspace{-5mm}

\end{table}

%% file: table/complexaction.tex
\begin{table}[t]
\centering

\resizebox{\columnwidth}{!}{
\begin{tabular}{lllc}
\toprule[1.5pt]
\textbf{Agent}         & \textbf{Input}              & \textbf{Base Model} & {\begin{tabular}[c]{@{}c@{}}\textbf{Action Success} \\ \textbf{Rate (\%)}\end{tabular}} \\
\midrule
\textit{Human}            & \textit{screen}             & \textit{-}     & \textit{100} \\

SeeAct \cite{seeact}               & SoM   & GPT-4 Turbo  &61.2  \\
AppAgent \cite{appagent}               & SoM   & GPT-4o & 53.2 \\
T3A \cite{androidworld}              & a11y tree             & GPT-4o     &59.7  \\
M3A \cite{androidworld}            & SoM             & GPT-4o     &64.5 \\
OS-Atlas \cite{osatlas}             & screen             & GPT-4o     &58.0  \\
UGround \cite{uground}     & screen             & GPT-4o     & 54.8\\
Mirage-O \cite{mirage}             & screen              & GPT-4o & 58.0 \\

\rowcolor[HTML]{DDDDDD}
RecAgen (Ours)   &a11y tree &GPT-4o & 64.5\\ 
\rowcolor[HTML]{DDDDDD}
RecAgen (Ours)   &SoM &GPT-4o & \textbf{69.3}\\ 
\bottomrule[1.5pt]
\end{tabular}
}
\vspace{-2mm}

\caption{Performance comparison on ComplexAction.}
\vspace{-5mm}

\label{tab:ca}
\end{table}

%% file: table/ablation1.tex
\begin{table}[t]
\centering
\begin{tabular}{ccc}
\hline
\multicolumn{2}{c}{Ablation Setting} & AndroidWorld \\ \hline
CRM     & \multicolumn{1}{c|}{RM}    & Success Rate (\%)          \\ \hline
\ding{55}        & \multicolumn{1}{c|}{\ding{55}}      &  40.5            \\
\ding{55}        & \multicolumn{1}{c|}{\ding{51}}      &  42.1            \\
\ding{51}        & \multicolumn{1}{c|}{\ding{55}}      &  43.5            \\
\ding{51}        & \multicolumn{1}{c|}{\ding{51}}      &   47.8           \\ \hline
\end{tabular}
\vspace{-2mm}

\caption{Ablation study on component recommendation module (CRM) and retrospection mechanism (RM).}
\vspace{-3mm}

\label{tab:ablation1}
\end{table}

%% file: table/ablation2.tex
\begin{table}[t]
\centering
\begin{tabular}{cccc}
\hline
\multicolumn{3}{c}{Recommendation Pathways} & ComplexAction \\ \hline
KMP & SMP & \multicolumn{1}{c|}{LRP} & Success Rate (\%)          \\ \hline
\ding{55}    & \ding{55}   & \multicolumn{1}{c|}{\ding{55}}  &   64.5           \\
 \ding{51}   & \ding{55}   & \multicolumn{1}{c|}{\ding{55}}  &    53.2         \\
\ding{55}    & \ding{51}   & \multicolumn{1}{c|}{\ding{55}}  &  56.4            \\
 \ding{55}   &  \ding{55}  & \multicolumn{1}{c|}{\ding{51}}  &    66.1          \\
 \ding{51}   &  \ding{51}  & \multicolumn{1}{c|}{\ding{51}}  &     69.3         \\ \hline
\end{tabular}
\vspace{-2mm}

\caption{Ablation study on different recommendation pathways. KMP, SMP and LRP indicate keyword matching pathway, semantic matching pathway and LLM-based intent recommendation pathway, respectively. When none of the three are used, it indicates that the component recommendation module is not employed.}
\vspace{-5mm}

\label{tab:ablation2}
\end{table}